\pdfoutput=1

\documentclass[11pt]{article}

\usepackage[final]{acl}

\usepackage{times}
\usepackage{latexsym}

\usepackage[T1]{fontenc}
\usepackage[utf8]{inputenc}
\usepackage{microtype}
\usepackage{inconsolata}
\usepackage{graphicx}

\usepackage{subfig}
\usepackage{float}
\usepackage{tikz}
\usepackage{amsmath}
\usepackage{amsfonts}
\usepackage{xcolor}
\usepackage{tcolorbox}
\usepackage{listings}
\tcbset{
colback=gray!10,
colframe=black,
fonttitle=\bfseries,
coltitle=white,
boxrule=0.8pt,
arc=3pt,
left=4pt,
right=4pt,
top=2pt,
bottom=2pt
}
\usepackage{cprotect}
\usepackage[capitalize]{cleveref}

\title{Using Source-Side Confidence Estimation for Reliable Translation into Unfamiliar Languages}

\author{Kenneth J. Sible \\
  University of Notre Dame \\
  \texttt{ksible@nd.edu} \\ \And
  David Chiang \\
  University of Notre Dame \\
  \texttt{dchiang@nd.edu} \\}

\begin{document}
\maketitle
\begin{abstract}
We present an interactive machine translation (MT) system designed for users who are not proficient in the target language. It aims to improve trustworthiness and explainability by identifying potentially mistranslated words and allowing the user to intervene to correct mistranslations. However, confidence estimation in machine translation has traditionally focused on the target side. Whereas the conventional approach to source-side confidence estimation would have been to project target word probabilities to the source side via word alignments, we propose a direct, alignment-free approach that measures how sensitive the target word probabilities are to changes in the source embeddings. Experimental results show that our method outperforms traditional alignment-based methods at detection of mistranslations.
\end{abstract}

\section{Introduction}
The demand for trustworthy and explainable AI has increased 
as models have become more and more complex. 
In machine translation (MT), one solution that has been around for decades is confidence estimation, which aims to assign a score to each sentence or each word indicating how reliable it is \citep{blatz_confidence_2004}. However, these methods have mainly focused on the target side,
enabling the user to post-edit the words or sentences that the MT system is least confident about. This is useful in applications where users are proficient in the target language but not the source language. 

But equally important are applications where users are proficient in the source language but not the target language. 
For example, a tourist traveling in a foreign country may want to ask a local if a certain food contains an allergen, so they decide to use an MT system to formulate their question in the local language. But they also want to be sure that what they say conveys their meaning correctly. In this situation, a strategy such as back-translation can be used to check the accuracy of the MT output. However, if back-translation reveals translation errors, the user has no straightforward way to guide the MT system toward a correct translation.

Although a user in such a situation may not be able to edit the MT output, they are able to edit the source text.
In this paper, we present an interactive MT system designed for this scenario. It uses source-side confidence estimation to highlight words with high uncertainty and provides the user with suggestions for source-side edits to improve translation quality.

\begin{figure}[t]
    \centering
    \begin{tikzpicture}
    \sbox0{\includegraphics[width=0.7\linewidth]{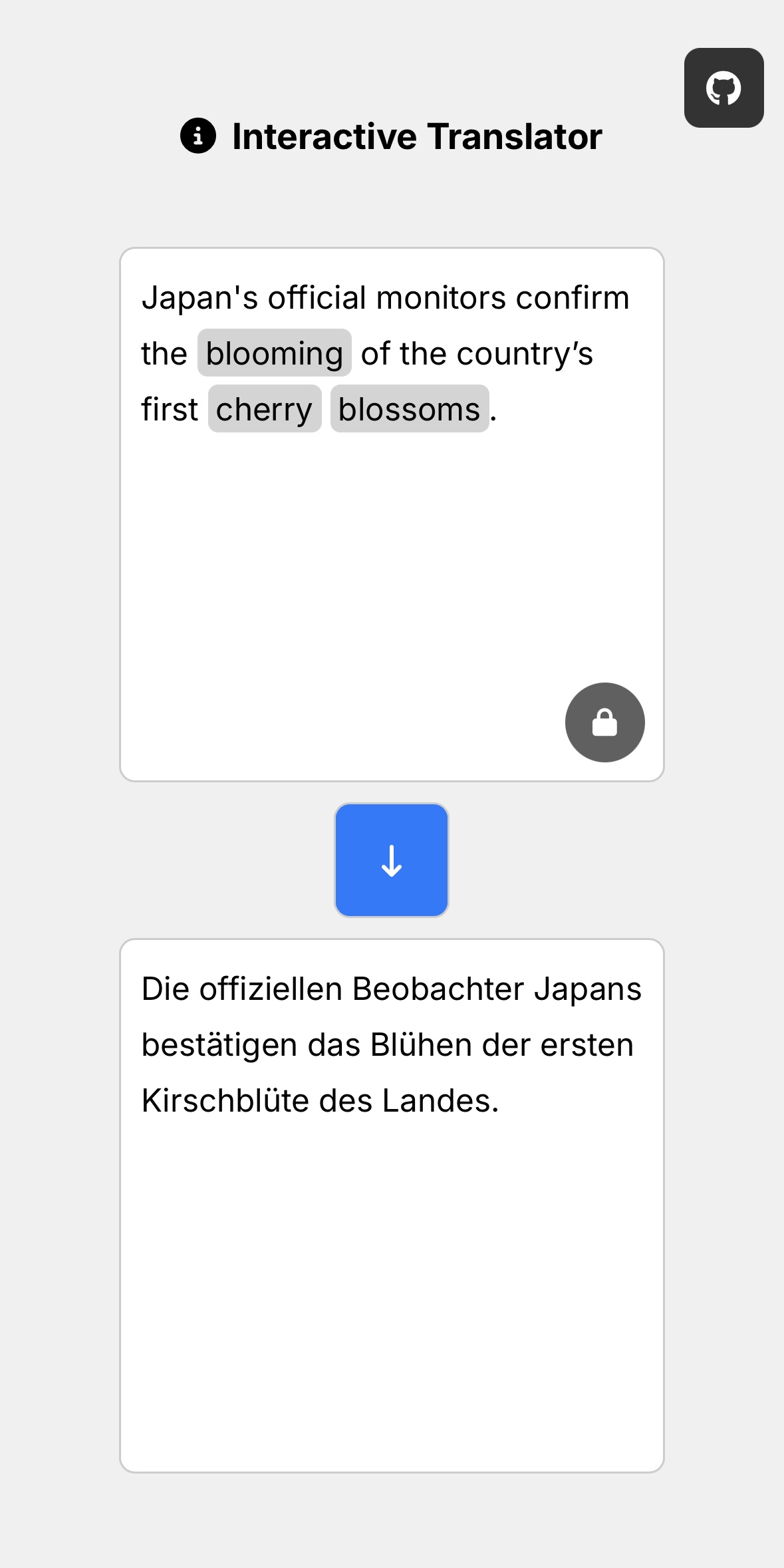}}%
    \path[clip,draw,rounded corners=.2cm] (0,0) rectangle (\wd0,\ht0);
    \path (0.5\wd0,0.5\ht0) node[inner sep=0pt]{\usebox0};
    \end{tikzpicture}
    \caption{A web application for an interactive MT system that highlights potentially mistranslated words that have been assigned high uncertainty scores by our gradient-based attribution method.}
    \label{fig:mobile-app}
\end{figure}

\Cref{fig:mobile-app} shows a screenshot of the mobile version of our system.
It currently translates English to German, and performs word-level confidence estimation on the source side, highlighting the uncertain words. If the user selects one of these words, it suggests alternative words/phrases. 

Our approach to source-side confidence estimation uses the gradient of the probability of the output sequence with respect to the source embeddings. In deep learning, this is known as an attribution method, which is a function that scores input features based on their contribution to the output of a model \citep{selvarajuGradCAMVisualExplanations2017,sundararajanAxiomaticAttributionDeep2017a}. For confidence estimation, we can interpret these scores as capturing how sensitive the output sequence is to the choice of source words, or rather to perturbations in the source embeddings.

Uncertainty scores could also be integrated into other applications, such as retrieving dictionary definitions for high-uncertainty words, as in \citet{sibleImprovingRareWord2024}'s approach for encoder-decoder models that uses attention masking to link rare words with their dictionary definitions.

\section{Related Work}
\Citet{niehuesModelingConfidenceSequencetoSequence2019} use a probability-based approach to confidence estimation, where the probabilities from the decoder output are treated as word-level confidence scores on the target side. To transform these into confidence scores on the source side, the authors developed an internal alignment model, which they claim improves performance over external alignment models. However, we were unable to reimplement their internal model and instead decided to use MGIZA\footnote{\url{https://github.com/moses-smt/mgiza}} \cite{gao_parallel_2008} for confidence projection.

\begin{figure*}[t]
    \centering\small
    \begin{tcolorbox}[title=LLM Prompt: Detecting Single-Word Translation Errors]
\textbf{Task:}  

Identify and list all single-word translation errors in a machine-translated sentence.

\vspace{1em}
\textbf{Instructions:}  

1. \textbf{Identify Mistranslations:} Compare each word in the candidate sentence with the source and reference sentences to find clear translation errors. 

2. \textbf{Focus on Single-Word Errors:} Report errors at the word level. If a multi-word phrase is mistranslated, extract the most significant word.  

3. \textbf{Prioritize Meaningful Errors:} Report only errors that significantly change meaning. Ignore acceptable variations, such as near-synonyms.  

4. \textbf{Ensure Contextual Accuracy:} Identify words that, while potentially valid in isolation, do not fit the intended meaning in context.  

5. \textbf{Handle Ambiguity Carefully:} Mark an error only if the candidate word is demonstrably incorrect when compared to the source and reference.  

\textbf{Strict Adherence Required:} Always follow the output format exactly, and include detailed explanations that refer back to the instructions.

\vspace{1em}
\textbf{Output Format:}  

List each mistranslation as a triple in the following format:  

\texttt{(source word → candidate word → reference word)}  

If there are no translation errors, output: 

\texttt{No translation errors detected.}

\vspace{1em}
\textbf{Example 1:}

\vspace{1em}
\textbf{Input:}

\textbf{Source Language:} English

\textbf{Target Language:} German

\textbf{Source Sentence:} The scientist presented a groundbreaking discovery.

\textbf{Candidate Translation:} Der Wissenschaftler präsentierte eine einfache Entdeckung.

\textbf{Reference Translation:} Der Wissenschaftler präsentierte eine bahnbrechende Entdeckung.

\vspace{1em}
\textbf{Expected Output:}  

\texttt{groundbreaking → einfache → bahnbrechende}  

\textbf{Explanation:} "einfache" (simple) is a mistranslation of "groundbreaking", which significantly changes the meaning of the sentence. This violates the \textbf{Prioritize Meaningful Errors} rule because it downplays the significance of the discovery.

    \end{tcolorbox}
    \cprotect\caption{A GPT-4o mistranslation detection prompt. The prompt is written to detect mistranslation pairs between a source sentence and a candidate sentence, and to match each pair with a word from the provided reference translation. In addition to these instructions, the prompt also contains some examples with explanations.}
    \label{fig:gpt-4o-prompt}
\end{figure*}

\begin{figure*}[t]
    \centering \small
    \begin{tcolorbox}[title=Example LLM Output: Detected Mistranslations]
\textbf{Input:}  

\textbf{Source Language:} English  

\textbf{Target Language:} German  

\textbf{Source Sentence:}  
\textcolor{blue!60!black}{Greenland sharks} swim \textcolor{blue!60!black}{leisurely} along the sea floor of the North Atlantic, covering an average of 1,220 metres in an hour.  

\textbf{Candidate Translation:}  
\textcolor{blue!60!black}{Grönland} schwimmt \textcolor{blue!60!black}{rasch} entlang des Meeresbodens des Nordatlantiks und umfasst durchschnittlich 1 220 m pro Stunde.  

\textbf{Reference Translation:}  
\textcolor{blue!60!black}{Gemächlich} schwimmt der \textcolor{blue!60!black}{Grönlandhai} am Grund des Nordatlantiks entlang, in einer Stunde kommt er im Durchschnitt gerade 1.220 Meter weit.  

\vspace{1em}
\textbf{Output:}  

\texttt{Greenland sharks → Grönland → Grönlandhai} 

\texttt{leisurely → rasch → Gemächlich}  

\textbf{Explanation:}  
"Grönland" (Greenland) is incorrectly used for "Greenland sharks", which should be "Grönlandhai". This is a significant translation error as it changes the subject of the sentence from the sharks to the island. Similarly, "rasch" (quickly) is a mistranslation of "leisurely", significantly altering the meaning of the sharks' swimming behavior. This violates the \textbf{Ensure Contextual Accuracy} rule.

    \end{tcolorbox}
    \cprotect\caption{A German source sentence from the test set along with the reference sentence, the MT candidate translation, and the GPT-4o output for the mistranslation detection task. The MT output translates back to "Greenland swims rapidly along the seafloor of the North Atlantic and covers an average of 1,220 meters per hour."}
    \label{fig:gpt-4o-output}
\end{figure*}

\section{Methodology}
In this section, we describe our gradient-based attribution approach to source-side confidence estimation, as well as a procedure for evaluating confidence estimation using LLM annotators.

\subsection{Gradient-Based Confidence Estimation}
For each source word $x_i$, we estimate uncertainty $\mathcal{U}(x_i)$ by measuring the sensitivity of the output sequence to the given source word. To compute sensitivity, we take the probability of the output sequence given the source sequence, $\mathbb{P}(y_1,\ldots,y_m\mid x_1,\ldots,x_n)$, and calculate the gradient with respect to the embedding vector of the given source word, $\mathbf{x}_i$. Finally, $\mathcal{U}(x_i)$ is reduced to a single value by taking the $L_1$ norm:

\begin{equation}
    \mathcal{U}(x_i)=\sum_{k=1}^{|\mathbf{x}_i|}\left\lvert\frac{\partial\mathbb{P}(y_1,\ldots,y_m\mid x_1,\ldots,x_n)}{\partial\mathbf{x}_i^k}\right\rvert.
    \label{eq:uncertainty}
\end{equation}

For brevity, we use $\mathcal{U}(x_i)$ to denote uncertainty, but $\mathcal{U}_i(x_1,\ldots,x_n)$ would be more appropriate since the uncertainty of $x_i$ depends on the context of the sentence. In other words, there may be a context in which the model can correctly translate the source word $x_i$ and another where it cannot.

If $\mathcal{U}(x_i)$ is small (low uncertainty), we say that the model has high confidence in its ability to translate the source word $x_i$ in the given context. This is because a small value of $\mathcal{U}(x_i)$ means that small changes in the embedding $\mathbf{x}_i$ have a small effect on the output sequence, that is, the model is robust to perturbations in $\mathbf{x}_i$. Similarly, a large value of $\mathcal{U}(x_i)$ (high uncertainty) corresponds to low confidence, since this means that small changes in $\mathbf{x}_i$ have a large effect on the output sequence.

To classify words as low/high uncertainty, we need to choose an uncertainty threshold to represent the decision boundary. Additionally, we also need to choose a subword aggregation function to compute uncertainty scores for entire words, since subword tokenization is commonly used in machine translation. For example, if $x_i$ consists of three subwords, we can sum their uncertainty scores to get a single score for the entire word. In Section~\ref{sec:experiments}, we describe the details of choosing an uncertainty threshold and an aggregation function.

\subsection{Evaluation Method: GPT-4o Annotator}
To evaluate our approach to confidence estimation, we translate a test set, annotate each mistranslation, and compute F1/AUC values. However, since this approach is model-dependent, any changes to our MT system or test set would require a reannotation, which is both expensive and time-consuming. Moreover, to compare with a previous approach to confidence estimation, we cannot use their annotated data. We want a low-cost, time-efficient annotator for consistent evaluation; an LLM such as GPT-4o is a good candidate for this.

In Figure~\ref{fig:gpt-4o-prompt}, we provide the prompt we used to instruct GPT-4o to detect mistranslations. Our prompt provides the LLM with the source sentence, a candidate translation from the MT system, and a reference translation from the test set. The LLM can use its own knowledge of the source and target languages to detect translation errors, but must use the reference translation to explain why a source word was labeled as a translation error. In addition, we use few-shot chain-of-thought prompting to enable in-context learning (for our output format) and encourage the LLM to justify its decisions.

By providing the LLM prompt and a model snapshot\footnote{gpt-4o-2024-11-20} that we used for annotation, we have devised an efficient and cost-effective way to compare future work. In Figure~\ref{fig:gpt-4o-output}, we show an example sentence pair from our test set, the machine-translated output, and the GPT-4o annotation. As we can see, GPT-4o correctly identified the translation errors, followed the prescribed output format, and provided an explanation that used the reference translation and referred back to the instructions.

\section{Experiments}
\label{sec:experiments}
In this section, we compare our gradient-based attribution approach with two baseline approaches that use the probabilities from the decoder output as confidence scores and project these scores to the source side using alignment methods.

For these two baseline approaches, we chose MGIZA and attention as the alignment methods. Although it has been shown that attention does not always correlate with word alignment, we still included a comparison with attention alignment because this approach allows uncertainty scores to be distributed across all source words compared to more traditional word alignment methods \citep{rikters_confidence_2017}. To compute the alignment weights, we averaged the attention values for all decoder layers and attention heads.

For dimension reduction, we tested Equation~\ref{eq:uncertainty} with $L_1$, $L_2$, and $L_\infty$ norms, and for subword aggregation, we tested $\mathrm{sum}(\cdot)$, $\mathrm{avg}(\cdot)$, and $\mathrm{max}(\cdot)$.

As evaluation metrics, we chose F1 and Area Under the Curve (AUC), both of which are suitable for binary classifiers. For mistranslation detection, a true positive is defined as a mistranslation, as indicated by GPT-4o, with high uncertainty, as indicated by our gradient-based attribution method. In our case, we have a heavy class imbalance because the number of true negatives (low uncertainty, correct translation) greatly exceeds the number of true positives (high uncertainty, mistranslation). 

However, this imbalance is acceptable for our task since recall is more important than precision for confidence estimation. Just because our model is uncertain about a given translation does not necessarily mean that the translation is wrong, but if there is a mistranslation, we want our method to detect it. A Precision-Recall (PR) curve visualizes the trade-off between precision and recall at different decision thresholds, where the positive class performance can be captured by AUC-PR.

We also provide a Receiver Operating Characteristic (ROC) curve that visualizes the trade-off between false positives (high uncertainty, correct translation) and false negatives (low uncertainty, mistranslation). An ROC curve considers the positive and negative classes equally, while AUC-ROC captures the overall class performance. Essentially, AUC-ROC measures the overall discriminative power of our gradient-based attribution method, but since we are more concerned with mistranslations with high uncertainty than we are with correct translations with low uncertainty, AUC-PR is a better performance metric.

\begin{figure*}[h]
    \centering
    \subfloat{\includegraphics[width=0.5\textwidth]{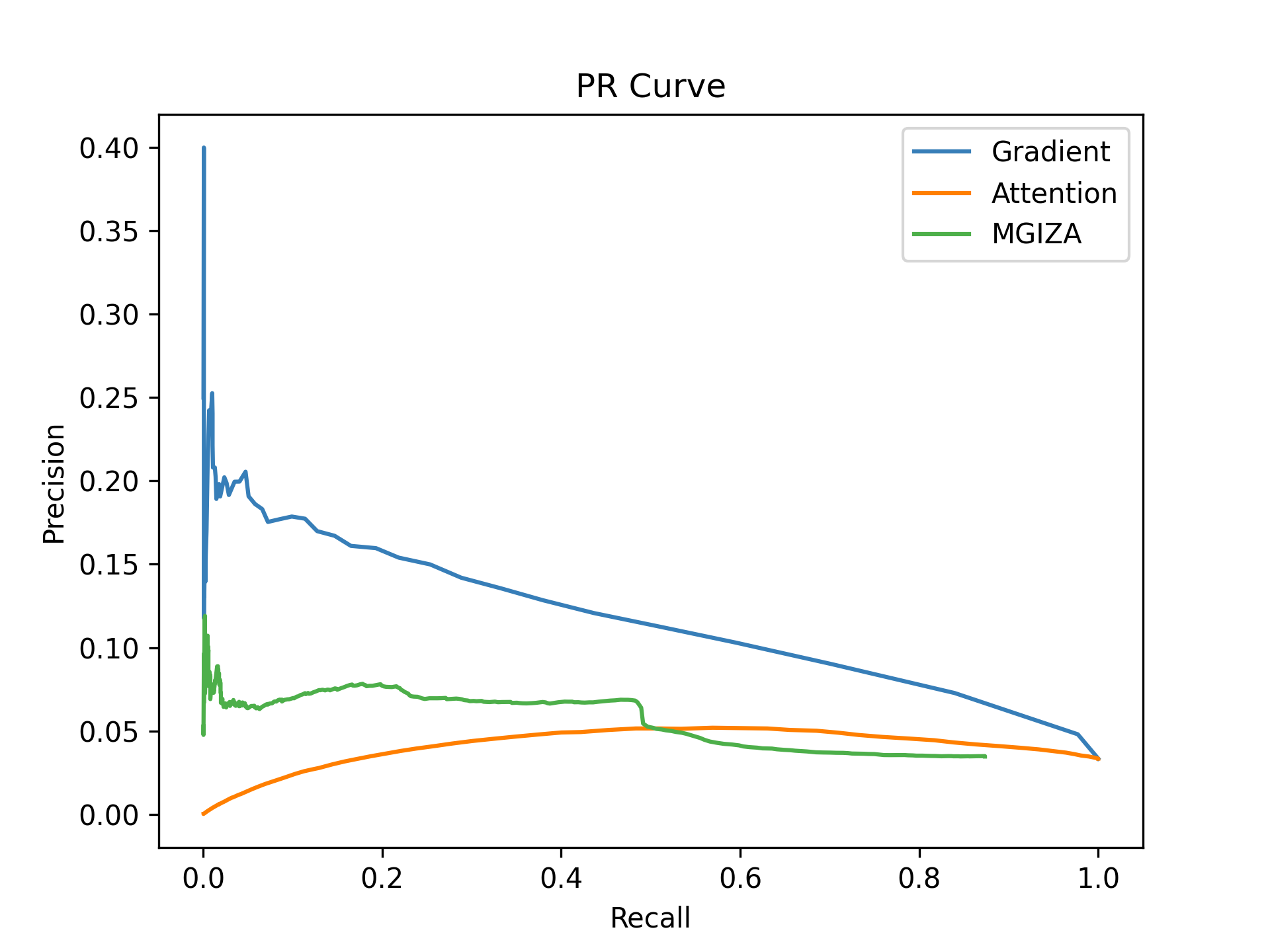}}
    \subfloat{\includegraphics[width=0.5\textwidth]{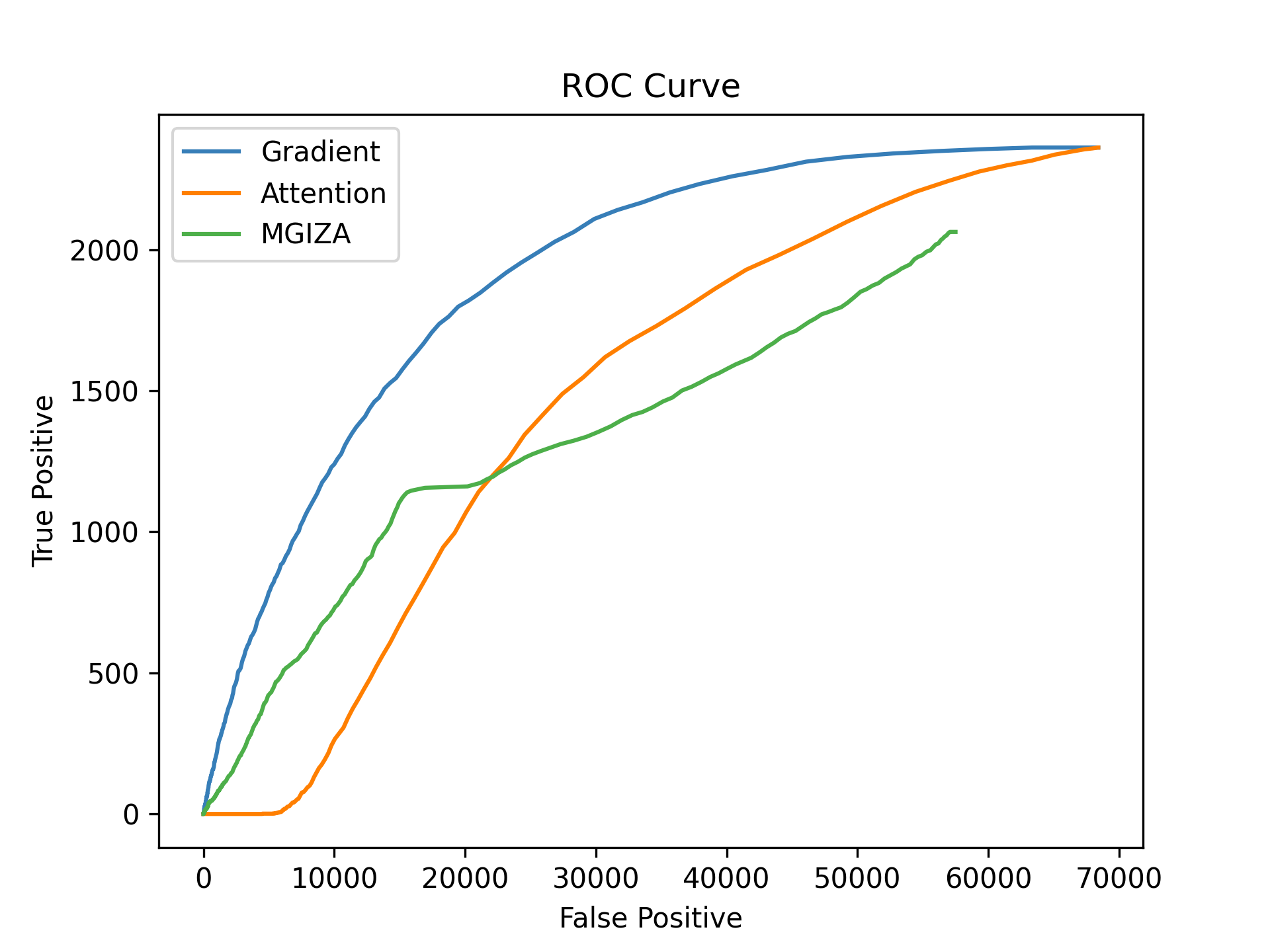}}
    \caption{Precision-Recall (PR) and Receiver Operating Characteristic (ROC) curves comparing our gradient-based attribution method with two baseline approaches for mistranslation detection. The PR curve (left) captures the performance of the positive class, which is more relevant for confidence estimation. The ROC curve (right) reflects the overall discriminative power of these methods as binary classifiers for mistranslation detection.}
    \label{fig:auc-curves}
\end{figure*}
\begin{table*}[h]
    \centering
    \begin{tabular}{r|c|c|c}
       \textbf{Model} & \textbf{Max F1} & \textbf{AUC-PR ($\times 10^{-2}$)} & \textbf{AUC-ROC ($\times 10^8$)} \\
       \hline
       MGIZA & 0.12 & 1.94 & 0.73 \\
       Attention & 0.10 & 0.77 & 1.00 \\
       Gradient & \textbf{0.19} & \textbf{8.36} & \textbf{1.31}
    \end{tabular}
    \caption{F1 scores and AUC values for the PR and ROC curves shown in Figure~\ref{fig:auc-curves}. AUC-PR is the greatest performance indicator for mistranslation detection, where the number of correct translations far exceeds the number of mistranslations. Higher F1 and AUC values indicate a greater ability to detect mistranslations.}
    \label{tbl:f1-scores}
\end{table*}

\section{Results}
Figure~\ref{fig:auc-curves} shows the PR and ROC curves for our gradient-based attribution method and the two baseline approaches using MGIZA and attention alignment, respectively. Table~\ref{tbl:f1-scores} shows the F1 and AUC values for the different methods, with the gradient-based attribution method performing best. We also found that $L_1$ is the optimal norm for dimension reduction and $\mathrm{sum}(\cdot)$ is the optimal function for subword aggregation.

\section{Applications}
This section describes a web application for an interactive MT system that highlights potentially mistranslated words and suggests alternatives. This interactive MT system is available in the browser and can be installed on a mobile device as a progressive web app (PWA),\footnote{\url{https://translate.kensible.com}} which is a standalone web app that can be added to the home screen.

\begin{figure*}[h]
    \begin{tikzpicture}
    \sbox0{\includegraphics[width=\linewidth]{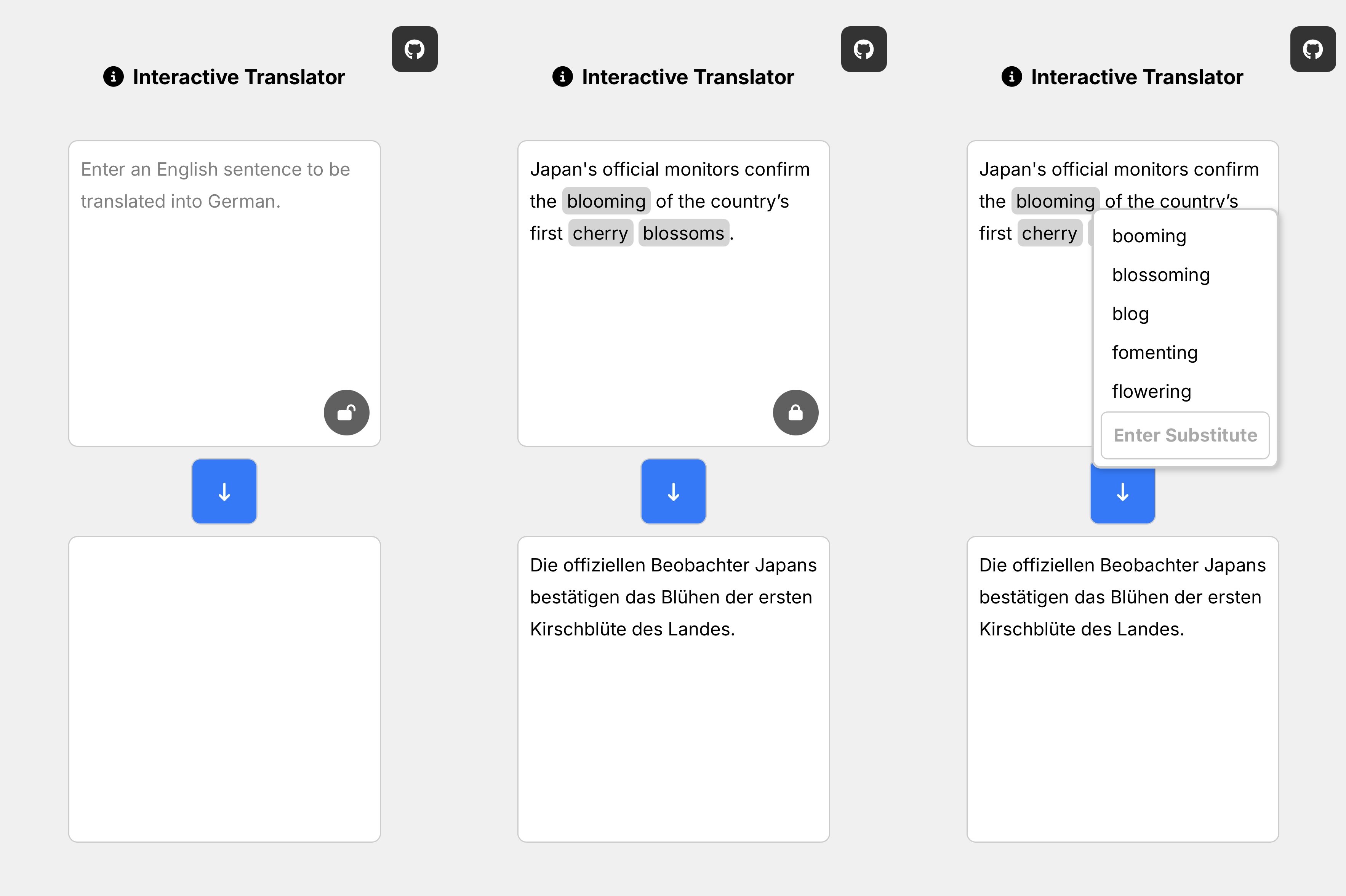}}%
    \path[clip,draw,rounded corners=.2cm] (0,0) rectangle (\wd0,\ht0);
    \path (0.5\wd0,0.5\ht0) node[inner sep=0pt]{\usebox0};
    \end{tikzpicture}
    \caption{\label{fig:side-by-side}An example of using the web application for an interactive MT system to translate a sentence from English to German. To start, the user enters an English sentence and clicks the translate button (left). When the translation appears, any potentially mistranslated words in the input are highlighted (middle). The user can then click on a highlighted word to see suggestions for alternative words, or enter their own replacement (right).}
\end{figure*}

Figure~\ref{fig:side-by-side} shows an example of using the interactive MT system in different stages. First, the user enters an English sentence to be translated into German. Here the user has entered "Japan's official monitors confirm the blooming of the country's first cherry blossoms." The words "blooming," "cherry," and "blossoms" are highlighted in the input field, indicating that these words have uncertainty scores below a decision threshold. We chose the threshold to be where the gradient-based attribution method achieved its maximum F1 score, which is 8.38.

As mentioned previously, the intended user of this web application is someone who is proficient in the source language but not the target language. In this case, if the user cannot understand German, they cannot tell that "blooming" was mistranslated to the verb "blühen" instead of the noun "Blüte." However, our gradient-based attribution method highlighted the word "blooming," so the user can select it to see suggestions that may improve the translation. Here we see five suggestions, including "blossoming" and "flowering," both of which are acceptable replacements in this context. Alternatively, the user can enter their own replacement if none of the suggestions are appropriate.

The suggested words are obtained from a $k$-nearest neighbors search using cosine similarity as the distance metric. The embeddings are the output of the final encoder layer in our MT system. Since we are using subword embeddings, these are averaged before indexing. We also prune words with a frequency of less than 10 from the vocabulary, since these low-frequency words tend to produce words that are orthographically similar, such as inflected forms or compound words.

To compute the nearest neighbors, we used Faiss,\footnote{\url{https://github.com/facebookresearch/faiss}} which is a library developed by Meta for efficient similarity search. Specifically, we used the inner product metric, \verb!faiss.IndexFlatIP!, with normalized embeddings. As for the MT system, we used our own custom PyTorch implementation, which is publicly available on GitHub.\footnote{\url{https://github.com/kennethsible/confidence-estimation}}

\section{Conclusion}
In this work, we introduced a gradient-based attribution approach to source-side confidence estimation for machine translation. Our method directly measures the sensitivity of the probability of the output sequence to perturbations in the source embeddings, eliminating the need for alignment-based projection. Experimental results show that our approach outperforms traditional alignment-based confidence estimation methods, allowing for more accurate detection of mistranslations.

In addition, we presented a web application for an interactive MT system that uses source-side confidence estimation to highlight potentially mistranslated words. This system is particularly beneficial for users who are proficient in the source language but unfamiliar with the target language, allowing them to refine their input sentence to achieve higher translation reliability.

To facilitate reproducibility and future comparisons, we proposed an evaluation framework using GPT-4o as an automatic annotator for mistranslation detection. By providing an LLM prompt and model snapshot, we establish a low-cost, scalable method for evaluating source-side confidence estimation. Our results show that GPT-4o produces consistent and reliable annotations, making it a valuable tool for evaluating new approaches.

We plan to integrate uncertainty scores into other applications, such as retrieving dictionary definitions for source words with high uncertainty. We also want to extend our web application to more language pairs and find better methods to generate semantically similar words, such as masked language modeling. We believe that our approach is a meaningful step towards more transparent and reliable machine translation systems.

\section*{Limitations}
Gradient-based attribution methods require backpropagation through the model, which increases the computational cost compared to simpler confidence estimation techniques. Additionally, specific model snapshots for LLMs may not be available indefinitely, potentially affecting the long-term reproducibility of our evaluation framework.

\section*{Acknowledgments}
This material is based on work supported by the National Science Foundation under Grant No. IIS-2137396.

\bibliography{custom}

\begin{thebibliography}{7}
\providecommand{\natexlab}[1]{#1}

\bibitem[{Blatz et~al.(2004)Blatz, Fitzgerald, Foster, Gandrabur, Goutte, Kulesza, Sanchis, and Ueffing}]{blatz_confidence_2004}
John Blatz, Erin Fitzgerald, George Foster, Simona Gandrabur, Cyril Goutte, Alex Kulesza, Alberto Sanchis, and Nicola Ueffing. 2004.
\newblock \href {https://aclanthology.org/C04-1046/} {Confidence {Estimation} for {Machine} {Translation}}.
\newblock In \emph{{COLING} 2004: {Proceedings} of the 20th {International} {Conference} on {Computational} {Linguistics}}, pages 315--321, Geneva, Switzerland. COLING.

\bibitem[{Gao and Vogel(2008)}]{gao_parallel_2008}
Qin Gao and Stephan Vogel. 2008.
\newblock \href {https://aclanthology.org/W08-0509/} {Parallel {Implementations} of {Word} {Alignment} {Tool}}.
\newblock In \emph{Software {Engineering}, {Testing}, and {Quality} {Assurance} for {Natural} {Language} {Processing}}, pages 49--57, Columbus, Ohio. Association for Computational Linguistics.

\bibitem[{Niehues and Pham(2019)}]{niehuesModelingConfidenceSequencetoSequence2019}
Jan Niehues and Ngoc-Quan Pham. 2019.
\newblock \href {https://doi.org/10.18653/v1/W19-8671} {Modeling {{Confidence}} in {{Sequence-to-Sequence Models}}}.
\newblock In \emph{Proceedings of the 12th {{International Conference}} on {{Natural Language Generation}}}, pages 575--583, Tokyo, Japan. Association for Computational Linguistics.

\bibitem[{Rikters and Fishel(2017)}]{rikters_confidence_2017}
Matīss Rikters and Mark Fishel. 2017.
\newblock \href {https://aclanthology.org/2017.mtsummit-papers.23/} {Confidence through {Attention}}.
\newblock In \emph{Proceedings of {Machine} {Translation} {Summit} {XVI}: {Research} {Track}}, pages 299--311, Nagoya Japan.

\bibitem[{Selvaraju et~al.(2017)Selvaraju, Cogswell, Das, Vedantam, Parikh, and Batra}]{selvarajuGradCAMVisualExplanations2017}
Ramprasaath~R. Selvaraju, Michael Cogswell, Abhishek Das, Ramakrishna Vedantam, Devi Parikh, and Dhruv Batra. 2017.
\newblock \href {https://doi.org/10.1109/ICCV.2017.74} {Grad-{{CAM}}: {{Visual Explanations}} from {{Deep Networks}} via {{Gradient-Based Localization}}}.
\newblock In \emph{2017 {{IEEE International Conference}} on {{Computer Vision}} ({{ICCV}})}, pages 618--626.

\bibitem[{Sible and Chiang(2024)}]{sibleImprovingRareWord2024}
Kenneth~J Sible and David Chiang. 2024.
\newblock Improving {{Rare Word Translation With Dictionaries}} and {{Attention Masking}}.
\newblock In \emph{Proceedings of the 16th {{Conference}} of the {{Association}} for {{Machine Translation}} in the {{Americas}} ({{Volume}} 1: {{Research Track}})}, pages 225--235, Chicago, USA. Association for Machine Translation in the Americas.

\bibitem[{Sundararajan et~al.(2017)Sundararajan, Taly, and Yan}]{sundararajanAxiomaticAttributionDeep2017a}
Mukund Sundararajan, Ankur Taly, and Qiqi Yan. 2017.
\newblock Axiomatic attribution for deep networks.
\newblock In \emph{Proceedings of the 34th {{International Conference}} on {{Machine Learning}} - {{Volume}} 70}, {{ICML}}'17, pages 3319--3328, Sydney, NSW, Australia. JMLR.org.

\end{thebibliography}

\end{document}